\documentclass{sig-alternate-2013}

\usepackage{graphicx}
\usepackage{mathtools}

\usepackage{amsmath,amssymb,amsfonts}

\usepackage{url}
\usepackage{verbatim}
\usepackage{multirow}
\usepackage{caption}
\usepackage{subcaption}

\usepackage[linesnumbered,ruled,vlined]{algorithm2e}
\usepackage{algpseudocode}
\usepackage{mathtools}

\usepackage{ifpdf}
\usepackage{lmodern}
\usepackage{lipsum}
\usepackage[T1]{fontenc}
\usepackage{textcomp}
\usepackage{enumitem}
\usepackage{mdwlist}
\usepackage{pbox}
\usepackage{comment}
\usepackage{ifthen}
\usepackage{color}
\usepackage{colortbl,xcolor}
\usepackage{upgreek}
\usepackage{upgreek}
\usepackage{todonotes}

\usepackage{tikz}
\usetikzlibrary{arrows}

\usepackage[hidelinks]{hyperref}

\definecolor{Gray}{gray}{0.9}
\definecolor{LightCyan}{rgb}{0.88,1,1}

\specialcomment{notes}{\begingroup \color{blue}} { \endgroup}

\newcolumntype{!}{>{\global\let\currentrowstyle\relax}}
\newcolumntype{^}{>{\currentrowstyle}}

\newcommand{\si}{\begin{enumerate}}

\newcommand{\ei}{\end{enumerate}}

\makeatletter
\let\oldfootnote\footnote
\def\footnote{\@ifstar\footnote@star\footnote@nostar}
\def\footnote@star#1{{\let\thefootnote\relax\footnotetext{#1}}}
\def\footnote@nostar{\oldfootnote}
\makeatother

\DeclareGraphicsExtensions{.eps,.png}
\graphicspath{{figure/}{figure/eps/}{figure/png/}}

\newcommand{\superscript}[1]{\ensuremath{^{\textrm{#1}}}}

\makeatletter
\def\@copyrightspace{\relax}
\makeatother

\begin{document}

\title{Actionable and Political Text Classification using Word Embeddings and LSTM}

\def\kloutinc{\superscript{*}}
\def\rutgers{\superscript{\dag}}

\numberofauthors{1} 
\author{
    \alignauthor Adithya Rao, Nemanja Spasojevic\\
    \affaddr{Lithium Technologies | Klout}  \\
    \affaddr{San Francisco, CA}  \\
    \email{\{adithya.rao, nemanja.spasojevic\}@lithium.com}
} 

\maketitle

\begin{abstract}
In this work, we apply word embeddings and neural networks with Long Short-Term Memory (LSTM) to text classification problems, where the classification criteria are decided by the context of the application.
We examine two applications in particular. 

The first is that of \textit{Actionability}, where we build models to classify social media messages from customers of service providers as \textit{Actionable} or \textit{Non-Actionable}.
We build models for over $30$ different languages for actionability, and most of the models achieve accuracy around $85$\%, with some reaching over $90$\% accuracy.
We also show that using LSTM neural networks with word embeddings vastly outperform traditional techniques.

Second, we explore classification of messages with respect to political leaning, where social media messages are classified as \textit{Democratic} or \textit{Republican}. 
The model is able to classify messages with a high accuracy of $87.57$\%.
As part of our experiments, we vary different hyperparameters of the neural networks, and report the effect of such variation on the accuracy.

These actionability models have been deployed to production and help company agents provide customer support by prioritizing which messages to respond to.
The model for political leaning has been opened and made available for wider use.
\end{abstract}

\category{J.4}{Computer Applications}{Social and Behavioral Sciences}
\category{H.1.2}{Information Systems}{Models and Principles}[User/Machine Systems]
\category{J.4}{Computer Applications}{Information Systems Applications}

\keywords{text classification; social media; deep learning; neural networks; lstm; actionability; politics;} 

\section{Introduction}
\label{section:introduction}
A large body of work in recent years in the field of text classification has explored sentiment mining \cite{socher2013recursive},  \cite{Socher2011}, \cite{glorot2011domain}, \cite{taboada2011lexicon}, \cite{agarwal2011sentiment}. 
In such problems, phrases or sentences are classified as positive or negative based on the sentiment expressed. 
While knowing the sentiment of a message certainly has its uses, other forms of useful text classification have remained relatively unexplored.
In this work explore scenarios where knowing only the sentiment of a piece of text is not useful enough to get good insights. 
We instead consider contextual classification of text, where the classification is based on criteria determined by the application.

One such application of contextual text classification is in the area of customer support. 
Public interactions on social networks provide an ideal platform for customers to interact with brands, companies and service providers. 
According to Okeleke \cite{ovum2014white} $50\%$ of users prefer reaching service providers on social media over contacting a call center.
A user may seek customer support from a service provider on social media with a complaint or a call for assistance.
In most cases where a customer may reach out in this manner, the message with the complaint expresses a negative sentiment. 
A service provider seeking to resolve such customer complaints may find little value in identifying that the sentiment of such messages is negative.

Instead, a better contextual classification is to classify the messages as \textit{Actionable} or \textit{Non-Actionable}. 
For such conversations, social media messages that include a clear call to action, or raise a specific issue can be categorized as \textit{Actionable}.
Alternatively, agents of the service provider may not be able to respond to \textit{Non-Actionable} messages that are too broad, general or not related to any specific issue.
The service providers could then prioritize responding to \textit{Actionable} messages over \textit{non-actionable} ones, saving money and resources in the process.
The ability to sift through interactions and classify them as actionable can help reduce company response latency and improve efficiency, thereby leading to better customer service.
A good solution to identify actionable messages can lead to large cost savings for companies by reducing call center loads. 

Another example where such contextual classification becomes important, is with respect to voter opinions during elections. 
During the election year in the United States, people on social media platforms such as Twitter may talk about various political issues, some leaning left with Democratic views and some leaning right with Republican views.
While expressing such views, a particular individual may speak positively about certain issues and negatively about others. 
In this scenario, merely understanding the sentiment of messages is not sufficient to provide enough insight into voter opinions.
Instead messages classified as \textit{Democratic} or \textit{Republican} may provide greater insight into the political preferences of different groups of people, and such classification may prove to be more valuable to a political candidate trying to understand his or her voters.

In recent studies, neural networks have shown great promise in performing sentiment and text classification tasks \cite{socher2013recursive}, \cite{Socher2011}, \cite{ykim2014}. 
Further, word embeddings have proven to be useful semantic feature extractors \cite{Mikolov2013word2vec}, \cite{glove2014}, and long-short term memory (LSTM) networks \cite{LSTM1997} have been shown to be quite effective in tasks involving text sequences \cite{sequence2sequence2014}.
Here we apply word embeddings and LSTM networks to the problem of contextual text classification, and present experiments and results for the two applications described above.

Our contributions in this study are as follows:
\begin{itemize}
  \item \textbf{Actionability:} We build models to classify messages as \textit{actionable} or \textit{non-actionable}, to help company agents providing customer support by prioritizing which messages to respond to. These models have been deployed to production.
  \item \textbf{Political Leaning:} We build a second set of models to predict political leaning from social media messages, and classify them as \textit{Democrat} or \textit{Republican}. We open this model and make it available for wider use.
  \item \textbf{RNN experiments:} We vary different hyperparameters of the RNN models used for classifying text in the applications above, and provide a report of the effect of each hyperparameter on the accuracy.
  \item \textbf{Multi-Lingual models:} We build models for over 30 different languages for actionability using the LSTM neural networks, and also compare these results with traditional machine learning techniques.
\end{itemize}




\section{Related Work}
\label{section:related}
Deep learning techniques have shown remarkable success in text processing problems.
The authors in \cite{ykim2014}, \cite{collobert2011natural} and \cite{zhang2015text} have applied Convolutional Neural Networks (CNNs) for a variety of NLP and text processing tasks with great results.
Recurrent Neural Networks (RNNs) have also shown to be very effective for text classification \cite{socher2013recursive}, and LSTM networks \cite{LSTM1997} in particular have shown to perform well for sequence based learning tasks \cite{sequence2sequence2014}.
Furthermore, word embeddings such as those in word2vec \cite{Mikolov2013word2vec} or GloVe \cite{glove2014}, map words and phrases to a lower dimensional space, which serve as semantic feature extractors that can be effectively used for training.
Given the inherently sequential nature of messages and sentences, we employ LSTM units in our neural networks in this work, with word embeddings for semantic feature extraction.

A significant body of work regarding the text classification has focused on the sentiment analysis. 
A wide range of research has been carried out, including the use of part-of-speech tagging \cite{agarwal2011sentiment}, lexicon approaches \cite{taboada2011lexicon}, rule-based \cite{chikersal2015sentu}, hybrid \cite{cambria2016affective}, \cite{cambria2014senticnet}, and deep learning \cite{glorot2011domain}, \cite{socher2013recursive},  \cite{Socher2011}.
A particular challenge for supervised learning for sentiment classification is that of human disagreements on labels, limiting models to have precision and recall values in the range of $0.75-0.82$ \cite{nigam2004towards}.
While the majority of research investigates text classification problems within a unified domain, it has been recognized that sentiment detection models may change depending on the domain context \cite{kanayama2006fully}, \cite{glorot2011domain}.
Here we further explore this idea of contextual text classification, and investigate classification of messages on social media based on criteria other than sentiment.

The first problem we consider is that of actionability of messages with respect to customer support on social media platforms.
Munro \cite{actionableHaiti} studied this problem of actionability in the context of the disaster relief, where subword and spatiotemporal models were used to classify short-form messages. 
Jin et. al. \cite{jin2013service} used KNN and SVM techniques to identify customer complaints on a dataset containing $5,500$ messages from an online Chinese hotel community.
Although the study showed promising results the dataset was limited and constrained to a very specific domain.
Our work here builds on our previous work in \cite{Spasojevic:actionability}, where manually extracted features and traditional supervised learning techniques were applied to this problem.

Another problem we consider is that of identifying political leaning of messages, by analyzing social media posts by users of such platforms.
A similar study was carried out by the authors in \cite{iyyer2014political}, where the datasets considered were derived from debate transcripts and articles by authors with well-known political leanings.
Here we instead analyze this problem for a dataset derived from social media messages, which are far more noisy in nature, and include colloquial language, abbreviations and slang. 
As noted by the authors in \cite{iyyer2014political}, for politics in the United States, a voter from either party may have a mixture of conservative or liberal views.
While the work in \cite{iyyer2014political} classified documents as \textit{liberal} or \textit{conservative}, here we classify messages along party lines, as \textit{Democratic} or \textit{Republican}.

\section{Methodology}
\label{section:system_overview}
Below, we walk through the pre-processing steps and the neural network architecture used for building our models.



\subsection{Pre-processing}

The input sample messages are tokenized, converting the message to a sequence of tokens which serves as the input to the neural network.
The tokens are also used to create a vocabulary of words to be used during training. 
The frequency of each token is used as the index representation for the token. 
The vocabulary to be used is then chosen as the top $V$ most frequent words seen. 
This vocabulary size also represents the maximum features considered in the input layer of the neural network. 

Each message is thus transformed into a sequence of indices corresponding to the tokens in the message. For example a phrase "the school" may be transformed to [1, 125]. If an out-of-vocabulary (OOV) token is encountered, then it is marked with a reserved index. 

We choose to use fixed length vectors during the training process.
We therefore select a sufficiently large length as the maximum permissible length for a message, and pad zero index values to a message when it is shorter than the maximum length. 

\begin{figure}[ht]
  \centering
  \fbox{\includegraphics[width=0.97\columnwidth]{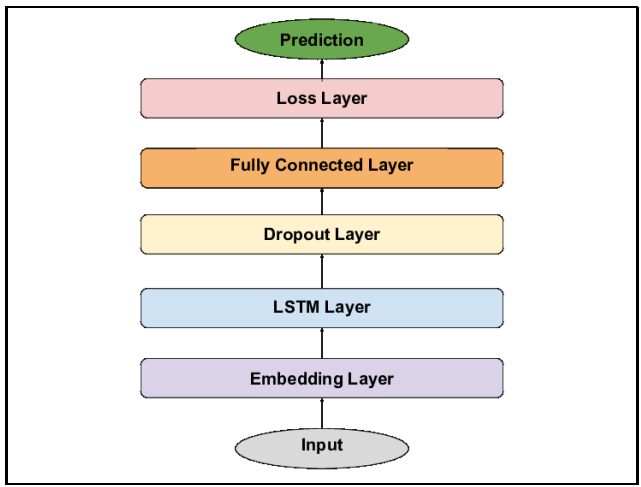}}
  \setlength{\abovecaptionskip}{0pt}
  \caption{Network Layers}
  \label{figure:network_layers}
  \vspace{-0.12in}
\end{figure}

\subsection{Neural Network Architecture}

Deep neural networks have shown success in a variety of machine learning tasks such as image recognition, speech processing and natural language processing. 
By using multiple processing layers with non-linear transformations, deep neural networks are able to model high-level abstractions in data. 
Recurrent Neural Networks (RNNs) are a class of neural networks that are able to model relationships in time. 
Unlike traditional neural networks, an RNN uses units with internal states that can persist information about previous events. 
This makes RNNs suitable for problems that require sequential information, such as text processing tasks.

Traditional RNNs suffer from the problem of not being able to learn long-term dependencies in data due to the problem of exploding and vanishing gradients. 
Long Short Term Memory (LSTM) units explicitly avoid this problem by regulating the information in a cell state using input, output and forget gates \cite{LSTM1997}. 
Such long term dependencies become an important consideration when learning to classify messages. 

The neural network used here employs word embeddings with LSTM units to perform contextual text classification.
We feed the pre-processed input to the neural network, and use the labels to perform supervised learning on the messages.
Figure \ref{figure:network_layers} shows these layers pictorially, and we describe each layer below.

\subsubsection{Embedding Layer}

The first layer in the network is an embedding layer that turns positive integer indices in the input into dense real valued vectors of fixed size, determined by the number of units in this layer. 
As a toy example, an input such as [[0], [4], [20]] may transformed to [[0.0, 0.0], [0.25, 0.1], [0.6, -0.2]].
The purpose of the embedding layer is to learn a mapping that embeds each word in the discrete vocabulary to a lower dimensional continuous vector space.
Such a distributed representation of vocabulary words has been shown to have great advantages in extracting relationships between word concepts in \cite{Mikolov2013word2vec} and \cite{glove2014}. 
The use of this layer enables semantic feature extraction from the input, without manual definition of features such as those in \cite{agarwal2011sentiment} and \cite{Spasojevic:actionability}.
The output of this layer is fed as the input to the further layers in the network.

\subsubsection{LSTM Layer}

Here we use a layer with multiple LSTM units as the second layer in the network. 
An LSTM unit is a memory cell composed of four main components: an input gate, a self-recurrent connection, a forget gate and an output gate. The input gate either blocks or allows an incoming signal to alter the cell-state. Similarly, the output gate either prevents or allows the cell-state from having han effect on other units. The forget gate allows the cell to remember or forget its previous state by controlling the cell's self-recurrent connection. 
An LSTM layer with multiple units can be thought of as a deep network across time steps, where each time step represents a layer.

In later sections we describe the effects of varying the number of units in the Embedding and LSTM layers.

\subsubsection{Dropout Layer}

Dropout \cite{Hinton2014Dropout} is a regularization technique used in neural networks to avoid overfitting. 
This is achieved by randomly dropping a fraction of the units while training a neural network, thus preventing units from co-adapting. 
The outputs of the LSTM layer in our network are fed to a Dropout layer where half the units dropout.

\subsubsection{Fully Connected Layer}

The fully connected layer has full connections to all the activations in the previous layer. 
This layer is essentially used to learn the non-linear combinations of the high-level features learnt by the previous layers in the network.

\subsubsection{Loss Layer}

The final layer in the network is the Loss layer that determines how deviation of the predicted labels from the actual labels are penalized.  
Since we are interested in binary classification of messages, here we use \textit{binary crossentropy} as the loss function.

Using this framework, we train our models for the two different applications of actionability and political leaning and evaluate the results in the following sections. 

\begin{table*}
\caption{\textbf{LSTM Language Models for Actionability.} ($V$: Vocabulary Size)} 
\centering
\setlength{\extrarowheight}{4pt}
\setlength{\arraycolsep}{5pt}
\begin{tabular}{|l||r|r|r|r|r|r|}
  \hline
  \rowcolor{Gray} \parbox[t]{1.5cm}{\textbf{Language}} & \parbox[t]{1.5cm}{\textbf{Training Samples}} & \parbox[t]{1.5cm}{\textbf{Test Samples}} & \parbox[t]{1.5cm}{\textbf{Training \newline Accuracy \newline ($V$=100k)}} & \parbox[t]{1.5cm}{\textbf{Test \newline Accuracy \newline ($V$=100k)}} & \parbox[t]{1.5cm}{\textbf{Training \newline Accuracy \newline ($V$=20k)}} & \parbox[t]{1.5cm}{\textbf{Test \newline Accuracy \newline ($V$=20k)}}\\
  \hline 
  af               & 174,119   & 43,530    & 0.8810 & 0.8525 & 0.8665 & 0.8547 \\ \hline
  \rowcolor{red!30}
  ar               & 380,038   & 95,010    & 0.8266 & 0.7940 & 0.8086 & 0.7950 \\ \hline
  \rowcolor{green!30}
  cs               & 31,407    & 7,852     & 0.9266 & 0.8988 & 0.9163 & 0.9087 \\ \hline
  da               & 134,623   & 33,656    & 0.9098 & 0.8844 & 0.8974 & 0.8905 \\ \hline
  de               & 223,438   & 55,860    & 0.8714 & 0.8316 & 0.8526 & 0.8352 \\ \hline
  en               & 6,841,344 & 1,710,337 & 0.8540 & 0.8508 & 0.8490 & 0.8475 \\ \hline
  es               & 720,783   & 180,196   & 0.8563 & 0.8406 & 0.8461 & 0.8402 \\ \hline
  et               & 113,635   & 28,409    & 0.9140 & 0.8874 & 0.9050 & 0.8958 \\ \hline
  \rowcolor{red!30}
  fa               & 21,991    & 5,498     & 0.8308 & 0.7641 & 0.8160 & 0.7643 \\ \hline
  \rowcolor{green!30}
  fi               & 93,844    & 23,462    & 0.9247 & 0.8951 & 0.9103 & 0.9013 \\ \hline
  fr               & 397,749   & 99,438    & 0.8349 & 0.8112 & 0.8197 & 0.8088 \\ \hline
  hr               & 48,728    & 12,182    & 0.9166 & 0.8747 & 0.8976 & 0.8771 \\ \hline
  hu               & 35,719    & 8,930     & 0.8251 & 0.8507 & 0.8814 & 0.8550 \\ \hline
  id               & 403,060   & 100,766   & 0.8826 & 0.8624 & 0.8699 & 0.8616 \\ \hline
  it               & 219,898   & 54,975    & 0.8915 & 0.8736 & 0.8818 & 0.8734 \\ \hline
  lt               & 22,257    & 5,565     & 0.8655 & 0.8762 & 0.8803 & 0.8652 \\ \hline
  nl               & 286,350   & 71,588    & 0.8530 & 0.8249 & 0.8361 & 0.8240 \\ \hline
  no               & 140,932   & 35,234    & 0.9056 & 0.8798 & 0.8944 & 0.8843 \\ \hline
  pl               & 79,386    & 19,847    & 0.9069 & 0.8668 & 0.8901 & 0.8787 \\ \hline
  pt               & 465,925   & 116,482   & 0.8545 & 0.8357 & 0.8441 & 0.8363 \\ \hline
  ro               & 81,920    & 20,481    & 0.9097 & 0.8726 & 0.8899 & 0.8798 \\ \hline
  \rowcolor{green!30}
  sk               & 73,808    & 18,452    & 0.9296 & 0.9000 & 0.9093 & 0.9098 \\ \hline
  sl               & 76,784    & 19,196    & 0.9120 & 0.8658 & 0.8878 & 0.8728 \\ \hline
  so               & 146,428   & 36,608    & 0.8911 & 0.8591 & 0.8773 & 0.8656 \\ \hline
  sq               & 31,683    & 7,921     & 0.8991 & 0.8794 & 0.8846 & 0.8799 \\ \hline
  \rowcolor{green!30}
  sv               & 114,395   & 28,599    & 0.9276 & 0.9013 & 0.9151 & 0.9069 \\ \hline
  sw               & 51,396    & 12,849    & 0.9029 & 0.8434 & 0.8778 & 0.8594 \\ \hline
  \rowcolor{red!30}
  th               & 52,859    & 13,215    & 0.8898 & 0.7601 & 0.8088 & 0.7997 \\ \hline
  tl               & 356,072   & 89,018    & 0.8487 & 0.8253 & 0.8374 & 0.8261 \\ \hline
  tr               & 123,217   & 30,805    & 0.8798 & 0.8291 & 0.8570 & 0.8331 \\ \hline
  vi               & 58,769    & 14,693    & 0.8769 & 0.8542 & 0.8694 & 0.8541 \\ \hline \hline
  \rowcolor{blue!30}
  Combined         & 1,417,723 & 354,431   & 0.9031 & 0.8597 & 0.8802 & 0.8681 \\ \hline 
\end{tabular}
\label{table:lstm_model_acc}
\end{table*}

\section{Actionability}
\label{section:actionability}

The first set of experiments is performed for the problem of actionability, where messages are classified as \textit{actionable} or \textit{non-actionable}.

\subsection{Datasets}

The dataset for this problem is derived from a social media management platform that helps agents respond to customer posts on behalf of their company.
The platform prioritizes incoming messages from customers and routes them to the appropriate agents, and is integrated with major social networks such as Twitter, Facebook and Google+, as well as with brand communities and online forums. 
This tool that is used by a number of brands and companies for customer support.

The labels for supervised training are gathered as follows.
The company agents using the platform may respond to certain incoming messages, and ignore those that they cannot provide a good response to.
If an agent provided a response, then the message is marked as \textit{Actionable} (labeled \textbf{1}), otherwise it is marked as \textit{Non-Actionable} (labeled \textbf{0}).
This labeled dataset is then used for training and evaluating models.

In this study we used a trailing window of 6 months of data, from November 1st 2014 to May 1st 2015 \footnote{All data analyzed here is publicly available on Twitter and Facebook and no private data was used as part of this research.}.
Equal number of actionable and non-actionable messages are sampled to create each dataset. 
The training and test datasets are created as 80/20 splits of the sampled data. 

We examine actionability across languages in our experiments.
A pair of training and test datasets is therefore created for each language, as well as one dataset that includes all languages. 
Some languages have sparse data and smaller number of samples, while others are larger. 
Our dataset sizes vary from 27k messages for the language with the least data (Farsi), to 8.5 million messages for the language with the most (English).
A combined language dataset is also prepared, which contains 1.7 million sampled messages across all languages. 
Table \ref{table:lstm_model_acc} shows the training and test dataset sizes for each language.




\subsection{Multi-lingual contexts}

Messages in different languages add an additional dimension to the problem of actionability.
We therefore generate separate models for over 30 different languages and compare their performance. 

Languages such as Czech (cs), Finnish (fi), Slovak (si) and  Swedish (sv) show the highest accuracies of over 90\%. 
The lowest accuracies are for languages such as Arabic (ar), Farsi (fa) and Thai (th), where the accuracies are slightly below 80\%, though in some cases like fa and th, the low accuracy may be attributed to sparsity of data.
As expected, certain languages are more predictable in terms of actionability than others, and we observe a variation of 15\% between the least and most accurate models.
Encouragingly, we find from Table \ref{table:lstm_model_acc} that most languages show high accuracies of greater than 85\%.

In addition, we also build a combined model trained on 1.7 million messages containing sampled data from all languages. 
Interestingly, this general model applicable across languages is able to capture actionability with a relatively high accuracy of 86\%. 
This suggests that the neural network can capture features across languages to identify actionability.

\subsection{Comparison with Traditional Methods}

In \cite{Spasojevic:actionability} we explored this problem through the lens of traditional supervised learning techniques, including logistic regression, AROW, perceptrons and Adagrad RDA.
Here we compare the results obtained using these traditional techniques, against the ones obtained using the LSTM network. 
In addition, in the previous study we extract features that are manually defined, whereas here we use word embeddings to extract semantic features.

We find from Table \ref{table:traditional_v_lstm} that the LSTM models consistently outperform the best traditional supervised models for almost all the languages, with only Arabic showing a slightly worse performance.
Further the improvement is significantly large with the LSTM model, the largest improvement being 15\% for Italian (it). 
This clearly demonstrates the superiority of neural networks when it comes to the text classification problems such as actionability.

\begin{table}[ht]
\small
\setlength\tabcolsep{4pt}
\caption{\textbf{Traditional vs LSTM Model Performance}}
\centering
\begin{tabular}{| l || r | r |}
  \hline
  \rowcolor{Gray} \textbf{Language} & \parbox{2.7cm}{\textbf{Traditional Model \newline Accuracy}} & \parbox{2.7cm}{\textbf{LSTM Model \newline Accuracy}} \\
  \hline
  \textbf{en} & 0.74 & 0.85 \\ \hline
  \textbf{es} & 0.76 & 0.84 \\ \hline
  \textbf{fr} & 0.76 & 0.81 \\ \hline
  \textbf{pt} & 0.71 & 0.84 \\ \hline
  \textbf{tl} & 0.79 & 0.83 \\ \hline
  \textbf{id} & 0.74 & 0.86 \\ \hline
  \textbf{af} & 0.75 & 0.85 \\ \hline
  \textbf{it} & 0.72 & 0.87 \\ \hline
  \textbf{nl} & 0.72 & 0.82 \\ \hline
  \textbf{ar} & 0.81 & 0.80 \\ \hline
\end{tabular}
\label{table:traditional_v_lstm}
\end{table}

\begin{table*}[!htp]
\setlength\tabcolsep{4pt}
\caption{\textbf{Non-Actionable vs Actionable Classification Examples}}
\centering
\setlength{\extrarowheight}{4pt}
\setlength{\arraycolsep}{5pt}
\begin{tabular}{| p{12cm} | r | l |}
  \hline
  \rowcolor{Gray} \textbf{Message} & \parbox{2cm}{\textbf{Prediction Score}} & \parbox{2cm}{\textbf{Classification}} \\
  \hline
\rowcolor{green!50}
@Verizon \#CEO confirms interest in possible @Yahoo acquisition & 0.02 & NON-ACTIONABLE \\ \hline
\rowcolor{green!40}
@PGE4Me's new program allows customers to go 100\% \#solar without having to install rooftop solar panels! & 0.14 & NON-ACTIONABLE \\ \hline 
\rowcolor{green!30}
Just another reason to despise @comcast and @xfinity Report: Comcast wants to limit Netflix binges & 0.15 & NON-ACTIONABLE \\ \hline
\rowcolor{green!20}
We're switching up the Official 1D @Spotify Playlist, which classic tracks do you want to see added? & 0.32 & NON-ACTIONABLE \\ \hline
\rowcolor{green!10}
I freaking HATE the changes @Yahoo has done to both the home page and my freaking mail!!! & 0.35 & NON-ACTIONABLE \\ \hline 
\rowcolor{orange!10}
Thanks for the 6 hour outage in my area @comcast ill def be calling and having money taken off my monthly bill. & 0.59 & ACTIONABLE \\ \hline
\rowcolor{orange!20}
@PGE4Me what's up with your online portal?  Taking forever to login and then everything says unavailable & 0.73 & ACTIONABLE \\ \hline
\rowcolor{orange!30}
@comcastcares @XFINITY Help! My wife just bought an episode of a show that we already watched (for free). Can I reverse the charge? & 0.82 & ACTIONABLE \\ \hline
\rowcolor{orange!40} 
@Yahoo I accidentally deleted my email, how do I get it back? & 0.85 & ACTIONABLE \\ \hline 
\rowcolor{orange!50}
@SpotifyCares hi someone has hacked my spotify and it won't let me listen to music because they are listening from another device help :( & 0.96 & ACTIONABLE \\ \hline
\end{tabular}
\label{table:actionability_examples}
\end{table*}

\subsection{Vocabulary size}

One of the parameters we consider while building models is the vocabulary size $V$. 
The vocabulary used for training the models in each language is chosen by picking the top $V$ most frequent words seen in the corpus. 
This parameter also determines the dimension of the feature space used for classification. 
A higher dimension could presumably capture more information, at the cost of larger models and slower training times.

In our experiments we consider two vocabulary sizes for each language. 
The first is a smaller vocabulary of size $20,000$, and the second is a larger one of size $100,000$.
The larger vocabulary adds significant overhead in training time, since more params need to be learnt by the model.
Table \ref{table:lstm_model_acc} shows the comparison of accuracies for the two vocabulary sizes.

In general, we observe that there is little difference between the test accuracy values for the two vocabulary sizes, with no decisive winner among them.
Thus even a smaller vocabulary is able to effectively capture actionability in messages, with the added advantage of faster training times. 

\subsection{Examples}

A useful side effect of the predictions made with the model, is that along with the final classification, the prediction score for a given message is indicative of how strongly actionable a message may be. 
Table \ref{table:actionability_examples} shows several examples of messages, ranging from least actionable messages with low scores, to most actionable ones with scores close to $1.0$.
We find that messages that simply convey some news are classified as the least actionable, and the ones which state an explicit issue with a call to action are classified as highly actionable.
Broad complaints or general questions are classified as non-actionable, or weakly actionable, which is a highly desired aspect when sifting through such messages.

These examples prove that the models are highly effective in classifying messages based on actionability.
Also note that there are many examples where a user may show a strong negative sentiment in the message, without the message being very actionable.
This again, highlights the need to classify messages on criteria other than sentiment alone.

\section{Political Leaning}
\label{section:political_leaning}





The second text classification problem we address is that of identifying political leaning with respect to politics in the United States. 
In this problem, we would like to classify messages as \textit{Democratic} or \textit{Republican}, based on the views expressed in the message.


\subsection{Dataset}

For this problem, we build the training and test datasets as follows:
First we pick users on Twitter whose political leanings are known to be either Democratic or Republican. 
We do this by using Twitter Lists, which are manually curated topical lists of users by other users of the social media platform.
For the users on these lists we sample the messages that they posted over a period of $3$ months, between Oct 12th, 2015 to Jan 12th, 2016.
The messages posted by the known Democrats as per the Twitter Lists are labeled \textbf{0}, and those by known Republicans are labeled \textbf{1}.
We then divide this dataset into the training and test sets with an 80/20 percent split.
The training set contains $336,000$ messages and the test set contains $84,000$ messages. 

\subsection{Training and Evaluation}

For training the model, we use a similar architecture as the one used for the actionability problem above, with 128 units each in the embedding and LSTM layers.
The model achieves a $88.82$\% accuracy on the training set, and a $87.57$\% accuracy on the test set.

We examine the effects of changing various parameters on the accuracy of the model.

\begin{figure*}
    \centering
    \begin{subfigure}[b]{0.32\textwidth}
        \includegraphics[width=\textwidth]{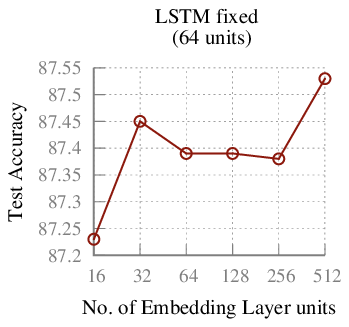}
        \caption{Accuracy vs Embedding Layer units}
        \label{fig:lstm_embed_lstm_64}
    \end{subfigure}
    ~ 
    \begin{subfigure}[b]{0.32\textwidth}
        \includegraphics[width=\textwidth]{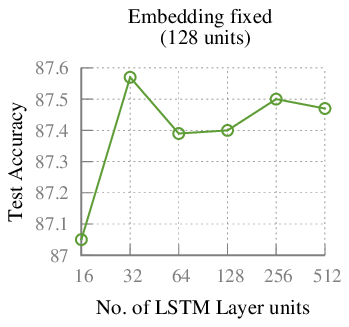}
        \caption{Accuracy vs LSTM Layer units}
        \label{fig:lstm_embed_128}
    \end{subfigure}
    ~ 
    \begin{subfigure}[b]{0.32\textwidth}
        \includegraphics[width=\textwidth]{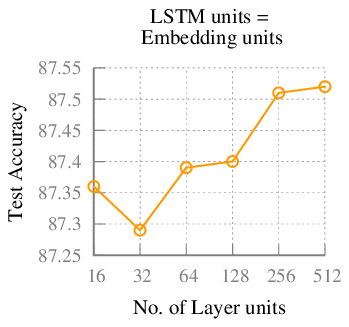}
        \caption{Accuracy vs Embedding/LSTM Layer units}
        \label{fig:lstm_equal_embed}
    \end{subfigure}
    \caption{Accuracy analysis as a function of number of LSTM and Embedding Layer Units.}
    \label{fig:accuracy_vs_units}
\end{figure*}

\begin{itemize}
\item \textbf{Embedding Layer Units}: For a fixed number of units in the LSTM layer ($64$), we observe that changing the number of units in the Embedding layer between $64$ to $256$ units shows very little change in accuracy.
When the units in the Embedding layer are lower ($32$) or higher ($512$), the accuracy is also higher, but if the number of units is too low ($16$) the accuracy drops. 
Figure \ref{fig:lstm_embed_lstm_64} shows this graphically. 

\item \textbf{LSTM Layer Units}: When the Embedding layer is held fixed at $128$ units, the highest accuracy is seen when the LSTM layer has $32$ units. Increasing the number of units beyond $32$ lowers the accuracy, and remains more or less constant thereafter. 
Figure \ref{fig:lstm_embed_128} shows the change in accuracy as a function of the LSTM layer units.

\item \textbf{Embedding and LSTM Layer Units}: When the Embedding and LSTM layers have the same number of units, larger sizes show higher accuracies, but the difference is nevertheless small. 
Figure \ref{fig:lstm_equal_embed} shows a distinct increasing trend in accuracy as the number of units are increased.

\item \textbf{Optimizers}: Among the tested optimizers, the 'Adam' optimizer \cite{kingma2014adam} performs the best with an accuracy of $87.57$\%, followed by 'Adagrad' with $87.12$\% and 'RMSprop' with $87.06$\%.

\item \textbf{Batch size}: A smaller batch size gives slightly higher accuracy, but the convergence is  faster for larger batch sizes. Based on our experiments, we pick a batch size of $64$, which gives a relatively high accuracy of $87.57$\% with a relatively fast training speed.

\item \textbf{Activation}: Finally, we observed the performance when the activation function in the last layer was changed. The sigmoid activation performed better than the tanh activation, with a difference of $0.5$\% in the accuracy.

\end{itemize}

\begin{table*}[ht]
\setlength\tabcolsep{4pt}
\caption{\textbf{Democrat vs Republican Classification Examples}}
\centering
\setlength{\extrarowheight}{4pt}
\setlength{\arraycolsep}{5pt}
\begin{tabular}{| p{12cm} | r | l |}
  \hline
  \rowcolor{Gray} \textbf{Message} & \parbox{2cm}{\textbf{Prediction Score}} & \parbox{2cm}{\textbf{Classification}} \\
  \hline
  \rowcolor{blue!50}
  The contrast between @HillaryClinton's depth of foreign policy knowledge and experience with this gang is astonishing. \#imwithher \#GOPDebate & 0.001 & DEMOCRATIC \\ \hline
  \rowcolor{blue!40}
  Climate change is real and caused by human activity. Our job is to aggressively transform our energy system away from fossil fuels. & 0.13 & DEMOCRATIC \\ \hline
  \rowcolor{blue!40}
  Both Hilary and Bernie mention climate change, the GOP...nothing. One of the biggest issues facing everyone, esp rural states who grow food. & 0.14 & DEMOCRATIC \\ \hline
  \rowcolor{blue!30}
  I like a lot of things about \#BernieSanders but one thing that stands out is that he doesn't talk down to the American ppl, esp millenials. & 0.20 & DEMOCRATIC \\ \hline
  \rowcolor{blue!30}
  @POTUS's FY 2017 Budget makes smart investments in tackling \#climatechange, accelerating innovation, making Americans safer. & 0.25 & DEMOCRATIC \\ \hline
  \rowcolor{blue!20}
  The decision to have an abortion is a right of any woman without interference from politicians. \#DemDebate & 0.31 & DEMOCRATIC \\ \hline
  \rowcolor{blue!20}
  Obama's super progressive budget would pour billions into clean energy, Medicaid, and edu while raising taxes & 0.41 & DEMOCRATIC \\ \hline
  \rowcolor{blue!10}
  Beyonce proudly displayed black power and they all collectively displayed LGBT love, the conservatives are seething  & 0.45 & DEMOCRATIC \\ \hline
  \rowcolor{red!10}
  Kansas oil producers are receiving less than \$25/barrel and Obama wants \$10 of it for his radical agenda. \#TCOT \#budget & 0.58 & REPUBLICAN \\ \hline
  \rowcolor{red!20}
  We will stop heroin and other drugs from coming into New Hampshire from our open southern border. We will build a WALL and have security. & 0.62 & REPUBLICAN \\ \hline
  \rowcolor{red!30}
  Is Obama clueless about Radical Islam? Nope he knows exactly what they are doing and he supports it \#RedNationRising  & 0.65 & REPUBLICAN \\ \hline
  \rowcolor{red!35}
  750 Billion has been paid out to illegal immigrants per Fox, under Obamacare subsidies! That is wrong!! Deport them all. Go Trump. & 0.66 & REPUBLICAN \\ \hline
  \rowcolor{red!45}
  THERE GOES FREEDOM: Twitter bans even MORE conservative accounts just for questioning Islam's attrocities! \#WakeUpAmerica \#1A & 0.69 & REPUBLICAN \\ \hline
  \rowcolor{red!50}
  RETWEET if you agree: Obama's \#MosqueVisit shows he underestimates the threat of radical Islam! & 0.75 & REPUBLICAN \\ \hline
  \rowcolor{red!60}
  PRAY FOR THE END OF ABORTION! Especially on \#AshWednesday and Everyday! \#Prolife & 0.94 & REPUBLICAN \\ \hline
  \rowcolor{red!60}
  Between Obama, ISIS, Congress, welfare leaches, unfettered immigration and Liberalism this country has gone downhill fast. & 0.97 & REPUBLICAN \\ \hline
\end{tabular}
\label{table:demvrep_examples}
\end{table*}





Figure \ref{fig:accuracy_vs_units} suggests that there may be an optimal number of Embedding and LSTM layer units that leads to the highest accuracy.
While we do see an increasing trend in accuracy when the Embedding and LSTM layers are simultaneously increased, but understanding how such hyperparameters should be chosen remains an open problem.
This line of inquiry requires further investigation into the factors that determine the optimal number of units in the Embedding and LSTM layers for a known input size. 
We note that since the changes in accuracy are relatively small overall, using a smaller network has advantages in terms of training times and memory constraints.


The model built with 128 units in the embedding and 32 units in the LSTM layer, trained with sigmoid activation and the Adam optimizer is available online for download \footnote{\url{https://github.com/klout/opendata}}. The list of users whose messages were aggregated, along with their known political leaning has also been made available.

\subsection{Examples}

As with actionability, the predictions for political leaning are accompanied by the prediction score, which indicate the extent of the leaning of the message.
Thus a message with a score of $0.9$ may express a strongly Republican view, whereas that with a score of $0.55$ may still be Republican, but may convey a more moderate view.
Table \ref{table:demvrep_examples} shows several examples of predictions made by the model.
We clearly see that the model is able to classify extreme views with very high or very low scores, and moderate views with scores closer to $0.5$.

Again, observe that Democratic as well as Republican messages both show positive and negative sentiments, irrespective of their leaning. 
Further, different users affiliated to the same party may have differing sentiments regarding certain issues.
This is another scenario where sentiment classification by itself may not provide enough insight into voter opinions, whereas classifying messages as \textit{Democratic} or \textit{Republican} may prove to be much more valuable.

\section{Conclusion}
\label{section:conclusion}
In this study, we explore two useful applications of text classification that go beyond sentiment analysis. 
For the purpose of such classification, we employ neural networks with word embedding and LSTM layers, and examine various parameters that affect the model.

The first application we consider is that in the context of customer support on social media, where company agents respond to customer complaints.
For this problem, we classify social media messages from customers as \textit{actionable} or \textit{non-actionable}, so that company agents can prioritize which messages to respond to.
We break down this problem and solve it for each language separately, and find that the models perform very well for most languages, with over 90\% accuracy for some of them.
We also find that the LSTM networks outperform traditional learning techniques by a large margin. 
Further a relatively small vocabulary size of $20,000$ and a training set of size $330,000$ is able to capture the features required to predict actionability. 

The second application we investigate is that of political leaning, where we classify messages as \textit{Democratic} or \textit{Republican}.
We try variations of different parameters of the neural networks and observe their effects on the accuracy.
Overall we see an accuracy of around 87\% in the best performing models for political leaning, and provide several illustrative examples of their effectiveness.

In the case of actionability most customer complaints carry a negative sentiment, whereas for political leaning both sides display a mixture of positive and negative sentiments.
As such, sentiment proves to be less valuable for such applications, since it is not a strong indicator of actionability or political leaning.
We therefore make the case here, that going beyond sentiment analysis to classify text based on other contextual criteria can lead to large benefits in various applications.

The actionability models described here are deployed in production and are used by paying customers, and the model for classifying political leaning has been made open and available for wider use.

\section{Acknowledgment}
\label{section:acknowledgement}
We would like to thank colleague David Gardner who was instrumental in deploying system actionability scoring service
in production.


\bibliographystyle{abbrv}
\bibliography{shared/latex/bibliography}

\begin{thebibliography}{10}

\bibitem{agarwal2011sentiment}
A.~Agarwal, B.~Xie, I.~Vovsha, O.~Rambow, and R.~Passonneau.
\newblock Sentiment analysis of twitter data.
\newblock In {\em Proceedings of the Workshop on Languages in Social Media},
  pages 30--38. Association for Computational Linguistics, 2011.

\bibitem{cambria2016affective}
E.~Cambria.
\newblock Affective computing and sentiment analysis.
\newblock {\em IEEE Intelligent Systems}, 31(2):102--107, 2016.

\bibitem{cambria2014senticnet}
E.~Cambria, D.~Olsher, and D.~Rajagopal.
\newblock Senticnet 3: a common and common-sense knowledge base for
  cognition-driven sentiment analysis.
\newblock In {\em Proceedings of the twenty-eighth AAAI conference on
  artificial intelligence}, pages 1515--1521. AAAI Press, 2014.

\bibitem{chikersal2015sentu}
P.~Chikersal, S.~Poria, and E.~Cambria.
\newblock Sentu: sentiment analysis of tweets by combining a rule-based
  classifier with supervised learning.
\newblock {\em SemEval-2015}, page 647, 2015.

\bibitem{collobert2011natural}
R.~Collobert, J.~Weston, L.~Bottou, M.~Karlen, K.~Kavukcuoglu, and P.~Kuksa.
\newblock Natural language processing (almost) from scratch.
\newblock {\em The Journal of Machine Learning Research}, 12:2493--2537, 2011.

\bibitem{glorot2011domain}
X.~Glorot, A.~Bordes, and Y.~Bengio.
\newblock Domain adaptation for large-scale sentiment classification: A deep
  learning approach.
\newblock In {\em Proceedings of the 28th International Conference on Machine
  Learning (ICML-11)}, pages 513--520, 2011.

\bibitem{LSTM1997}
S.~Hochreiter and J.~Schmidhuber.
\newblock Long short-term memory.
\newblock In {\em Neural Computation}, pages 1735--1780, 1997.

\bibitem{iyyer2014political}
M.~Iyyer, P.~Enns, J.~L. Boyd-Graber, and P.~Resnik.
\newblock Political ideology detection using recursive neural networks.
\newblock In {\em ACL (1)}, pages 1113--1122, 2014.

\bibitem{jin2013service}
J.~Jin, X.~Yan, Y.~Yu, and Y.~Li.
\newblock Service failure complaints identification in social media: A text
  classification approach.
\newblock 2013.

\bibitem{kanayama2006fully}
H.~Kanayama and T.~Nasukawa.
\newblock Fully automatic lexicon expansion for domain-oriented sentiment
  analysis.
\newblock In {\em Proceedings of the 2006 Conference on Empirical Methods in
  Natural Language Processing}, pages 355--363. Association for Computational
  Linguistics, 2006.

\bibitem{ykim2014}
Y.~Kim.
\newblock Convolutional neural networks for sentence classification.
\newblock {\em Proceedings of the 2014 Conference on Empirical Methods in
  Natural Language Processing (EMNLP)}, page 1746–1751, 2014.

\bibitem{kingma2014adam}
D.~Kingma and J.~Ba.
\newblock Adam: A method for stochastic optimization.
\newblock {\em arXiv preprint arXiv:1412.6980}, 2014.

\bibitem{Mikolov2013word2vec}
T.~Mikolov, I.~Sutskever, K.~Chen, G.~Corrado, and J.~Dean.
\newblock Distributed representations of words and phrases and their
  compositionality.
\newblock In {\em Proceedings of NIPS, 2013}, 2013.

\bibitem{actionableHaiti}
R.~Munro.
\newblock Subword and spatiotemporal models for identifying actionable
  information in haitian kreyol.
\newblock In {\em Proceedings of the Fifteenth Conference on Computational
  Natural Language Learning}, pages 68--77, 2011.

\bibitem{nigam2004towards}
K.~Nigam and M.~Hurst.
\newblock Towards a robust metric of opinion.
\newblock In {\em AAAI spring symposium on exploring attitude and affect in
  text}, pages 598--603, 2004.

\bibitem{ovum2014white}
A.~Okeleke and S.~Bali.
\newblock {Integrating social media into CRM for next generation customer
  experience}.
\newblock Technical report, Ovum, 05 2014.

\bibitem{glove2014}
J.~Pennington, R.~Socher, and C.~D. Manning.
\newblock Glove: Global vectors for word representation.
\newblock In {\em Proceedings of EMNLP, 2014}, page 1532–1543, 2014.

\bibitem{Socher2011}
R.~Socher, J.~Pennington, E.~H. Huang, A.~Y. Ng, and C.~D. Manning.
\newblock Semi-supervised recursive autoencoders for predicting sentiment
  distributions.
\newblock EMNLP '11, pages 151--161, 2011.

\bibitem{socher2013recursive}
R.~Socher, A.~Perelygin, J.~Y. Wu, J.~Chuang, C.~D. Manning, A.~Y. Ng, and
  C.~Potts.
\newblock Recursive deep models for semantic compositionality over a sentiment
  treebank.
\newblock In {\em Proceedings of EMNLP}, volume 1631, page 1642, 2013.

\bibitem{Spasojevic:actionability}
N.~Spasojevic and A.~Rao.
\newblock Identifying actionable messages on social media.
\newblock In {\em IEEE International Conference on Big Data}, IEEE BigData '15,
  2015.

\bibitem{Hinton2014Dropout}
N.~Srivastava, G.~Hinton, A.~Krizhevsky, I.~Sutskever, and R.~Salakhutdinov.
\newblock Dropout: A simple way to prevent neural networks from overfitting.
\newblock In {\em Journal of Machine Learning Research 15}, pages 1929--1958,
  2014.

\bibitem{sequence2sequence2014}
I.~Sutskever, O.~Vinyals, and Q.~V. Le.
\newblock Sequence to sequence learning with neural networks.
\newblock In Z.~Ghahramani, M.~Welling, C.~Cortes, N.~Lawrence, and
  K.~Weinberger, editors, {\em Advances in Neural Information Processing
  Systems 27}, pages 3104--3112. Curran Associates, Inc., 2014.

\bibitem{taboada2011lexicon}
M.~Taboada, J.~Brooke, M.~Tofiloski, K.~Voll, and M.~Stede.
\newblock Lexicon-based methods for sentiment analysis.
\newblock {\em Computational linguistics}, 37(2):267--307, 2011.

\bibitem{zhang2015text}
X.~Zhang and Y.~LeCun.
\newblock Text understanding from scratch.
\newblock {\em arXiv preprint arXiv:1502.01710}, 2015.

\end{thebibliography}

\balancecolumns
\end{document}